\documentclass[twocolumn]{article}

\usepackage[accepted]{arxiv}

\newif\ifarxiv
\arxivtrue
\newif\ifslow
\slowfalse

\usepackage{times}
\usepackage{url}
\let\ifpdf\relax
\usepackage{graphicx} 
\usepackage[utf8]{inputenc}

\usepackage{color}

\usepackage{natbib}

\usepackage{algorithm}
\usepackage{algorithmic}

\usepackage{amsmath}

\renewcommand{\vec}[1]{{\boldsymbol{#1}}}

\def\ss{\vspace*{-.25mm}}
\def\sa{\vspace*{-.5mm}}
\def\sb{\vspace*{-1mm}}
\def\sc{\vspace*{-1.5mm}}

\begin{document} 

\title{Early Inference in Energy-Based Models Approximates Back-Propagation}

\author{Yoshua Bengio, CIFAR Senior Fellow\\
  and Asja Fischer\\
Montreal Institute for Learning Algorithms, University of Montreal}

\date{}

\maketitle

\sb
\begin{abstract}
We show that Langevin MCMC inference in an energy-based model with latent variables has the property that the early steps of inference, starting from a stationary point, correspond to propagating error gradients into internal layers, similarly to back-propagation. The error that is back-propagated is with respect to visible units that have received an outside driving force pushing them away from the stationary point. Back-propagated error gradients correspond to temporal derivatives of the activation of hidden units. This observation could be an element of a theory for explaining how brains perform credit assignment in deep hierarchies as efficiently as back-propagation does. In this theory, the continuous-valued latent variables correspond to averaged voltage potential (across time, spikes, and possibly neurons in the same minicolumn), and neural computation corresponds to approximate inference and error back-propagation at the same time.
\end{abstract}

\sb
\section{Introduction}
\sb

It has been hypothesized numerous
times~\citep{Hinton-bo86,Friston+Stephan-2007,Berkes-et-al-2011}, that,
given a state of sensory information (current and past inputs), neurons are
collectively performing {\em inference}, i.e., moving towards
configurations that better ``explain'' the observed sensory data.  We can
think of the configuration of internal neurons (hidden units or latent
variables) as an ``explanation'' for the observed sensory data.

Under this hypothesis, when an unexpected signal arrives at visible
neurons\footnote{in contact with the outside world, sensory neurons or internal neurons
  that directly reflect the sensory
signal}, the other neurons will change their state (from one that is near a
stochastic equilibrium) so as to better reflect that change in input.  If this
error signal is slowly driving visible neurons towards the externally observed
value, this creates a perturbation for the whole network. In this paper,
we consider what happens early in this process, when the perturbation due to
the incorrect prediction of visible neurons is propagated towards inner
areas of the brain (hidden units and hidden layers). We show how the propagation
of this perturbation is
mathematically equivalent to the propagation of activations gradients by
the back-propagation algorithm in a deep neural network.

This result assumes that latent variables are continuous-valued
(unlike in the Boltzmann machine), making the system an energy-based model similar
to the continuous Hopfield net~\citep{Hopfield84},
defined by an energy function over both continuous visible and hidden units.
It also assumes that the neurons are leaky integrators with injected noise. As a consequence
neural computation corresponds to gradually going down the energy
while adding noise, which corresponds to running a Langevin Monte-Carlo Markov chain (MCMC)
as inference mechanism.

This work is only a stepping stone.
Several points still need to be elucidated before a complete theory of learning,
inference and credit assignment is elaborated that is both biologically
plausible and makes sense from a machine learning point of view for whole
networks (with global optimization of the whole network and not being
limited to learning of visible neurons that receive a target). In particular,
energy based models require symmetry of connections, but note that hidden
units in the model need not correspond exactly to actual neurons in the brain
(it could be groups of neurons in a cortical microcircuit, for example).
It remains to be shown how a form of symmetry could arise from the learning
procedure itself, for example because of an unsupervised learning objective
similar to the one used to train autoencoders.
As shown in this paper, synaptic changes approximately proportional to the
stochastic gradient with respect to the prediction error
could be achieved if the synaptic updates follow the learning rule
studied by~\citet{Bengio-et-al-arxiv2015} in
order to mimic spike-timing dependent plasticity (STDP).

\sa
\section{Neural computation does inference: going down the energy}
\label{sec:inference}
\sa

We consider the hypothesis that a central interpretation of neural computation
(what neurons do, on a short time scale at which weights can be considered fixed)
is that it is performing {\em iterative inference}. Iterative
inference means that the hidden units $\vec h$ of the network are gradually changed
towards configurations that are more probable, with the given sensory input $\vec x$ and
according to the current ``model of the world'' associated with the parameters of the
model. In other words, they are approximately moving towards configurations more
probable under $P(\vec h | \vec x)$, and eventually sampling from $P(\vec h | \vec x)$.

Before making the connection between Boltzmann machines or energy-based models
and back-propagation, let us see in more mathematical detail how neural
computation could be interpreted as inference.

\subsection{Leaky integrator neuron as Langevin MCMC}

For this purpose, consider the classical leaky integrator neural equation.
Let $\vec s_t$ represent the state of the system at time $t$, a vector with one element per unit,
where $s_{t,i}$ is a real valued quantity associated with the $i$-th unit, corresponding
to a time-integrated voltage potential.

Let us denote $\vec x_t$ for the visible units, a subset of the elements of $\vec s_t$ (i.e., the externally driven
input) and $\vec h_t$ for the hidden units, i.e., $\vec s_t = (\vec x_t, \vec h_t)$. Let $f$ be the function
that computes the new value of the complete state $\vec s_t$ given its previous value, with $f=(f_x,f_h)$ to denote
the parts of $f$ that respectively outputs the predictions on the clamped (visible) units and on the unclamped
(hidden) units. The time evolution of the unclamped units is assumed to follow a leaky integration equation,
i.e., 
\begin{equation}
\sa
\label{eq:s-update}
 \vec  h_{t+1} = f_h(\vec s_t, \vec \eta_t) = \vec h_t + \epsilon (R_h( \tilde{ \vec s}_t) - \vec h_t)
\ss
\end{equation}
where $R(\vec s)=(R_x( \vec s),R_h(\vec s))$ represents the network-generated pressure on neurons, i.e.,
$R_i(\vec s)$ is what the rest of the network asks neuron $i$ to move towards,
and the corrupted state $\tilde{\vec s}$ is results from synaptic noise and
spiking effects. Here we roughly model this corruption by simple additive noise:
\begin{equation}
\ss
\label{eq:tilde-s}
  \tilde{\vec s}_t = \vec s_t + \vec \eta_t \enspace,
\ss
\end{equation}
and we see that the above equation corresponds to the discretization of a differential equation
\[
 \tau \dot{\vec h} = R_h(\vec s+\vec \eta) - \vec h
\]
which brings $\vec h$ exponentially fast towards the ``target'' value $R_h(\vec s)$, along with
the random walk movements brought by the noise $\vec \eta$.  %
We assume that $R_i(\vec s)$ is a weighted sum of input signals coming from the neurons
connected into neuron $i$, although the derivation below does not depend
on the specific form of $R$, only on the assumption that it corresponds to
an energy gradient, an idea developed below.

\subsection{Machine Learning Interpretation}

$R_h(\tilde{\vec s}_t)$ in Eq.~\ref{eq:s-update}
represents a guess for a new configuration, with 
$R_h(\tilde{\vec s}_t)-\vec h_t$
a noisy direction of movement. A noise-free direction would be $R_h(\vec s_t)-\vec h_t$,
but injecting noise is important in order to find not just a
single local mode of $P(\vec h | \vec x)$ but explore the full distribution.

We now draw an interesting link with recent work
on unsupervised learning using denoising auto-encoders and denoising
score matching~\citep{Vincent-2011,Alain+Bengio-ICLR2013-small}. If
$R(\vec s)$ is the linear combination of input rates $\rho(\vec s)$, the above
papers make a link between $R(\vec s)-\vec s$ and the energy of a probabilistic
model $P(\vec s)\propto e^{-E(\vec s)}$ with energy function $E(\vec s)$, i.e., they find that 
\begin{equation}
\label{eq:R}
R(\vec s)-\vec s \propto \frac{\partial \log P(\vec s)}{\partial \vec s} = - \frac{\partial E(\vec s)}{\partial \vec s}\enspace .
\end{equation}
With this interpretation, the leaky integration neural 
computation of Eq.~\ref{eq:s-update} seems to follow a Langevin Monte-Carlo Markov chain~\citep{Andrieu2003}:
\begin{align}
\ss
  \vec s_{t+1} &= \vec s_{t} + \epsilon (R(\tilde{\vec s}_{t}) - \vec s_{t}) \nonumber \\
      &= \vec s_{t} + \epsilon(R(\tilde{\vec s}_{t})-\tilde{\vec s}_{t} + \tilde{\vec s}_{t} - \vec s_{t}) \nonumber \\
      &= \vec s_{t} + \epsilon(-\frac{\partial E(\tilde{\vec s}_{t})}{\partial \tilde{\vec s_{t}}} + \vec \eta_{t})
\sa
\end{align}
where for the last line we used Eqs.~\ref{eq:tilde-s} and~\ref{eq:R} and we see that we are
going down the gradient from $\tilde{\vec s}_{t}$. Hence from the point of view of the noisy states $\tilde{\vec s}$,
we see that the update equation corresponds to
\begin{align}
\ss
\label{eq:update-tilde-s}
  \tilde{\vec s}_{t+1} - \eta_{t+1} &= \tilde{\vec s}_{t} - \vec \eta_{t} - \epsilon \frac{\partial E(\tilde{\vec s}_{t})}{\partial \tilde{\vec s_{t}}} + \epsilon \vec \eta_{t} \nonumber \\
  \tilde{\vec s}_{t+1} &= \tilde{\vec s}_{t} - \epsilon \frac{\partial E(\tilde{\vec s}_{t})}{\partial \tilde{\vec s_{t}}} + \vec \eta_{t+1} -(1- \epsilon) \vec\eta_{t}
\ss
\end{align}
which we recognize as going down the gradient of the energy with ``learning rate'' $\epsilon$ 
and adding ``noise'' $\vec \eta_{t+1} -(1- \epsilon) \vec \eta_{t}$. 

\subsection{A Possible Energy Function}

To fix ideas and illustrate the possibility of a driving function $R$
corresponding to the gradient of an energy function, we propose the
following energy function, closely related to the Boltzmann machine
energy function, but with continuous non-linearities inserted.
\begin{equation}
\sa
\label{eq:energy-function}
 E(\vec s) = \sum_i \frac{s_i^2}{2} - \frac{1}{2}\sum_{i\neq j} W_{i,j} \rho(s_i) \rho(s_j) -\sum_i b_i \rho(s_i) \enspace
\ss
\end{equation}
where $W_{i,j}$ is the weight between unit $j$ to unit $i$, and
$\rho$ is the neural non-linearity, some kind of monotonic bounded function which outputs a value between 0 and 1 corresponding to a firing rate, as for example
\begin{equation}
\label{eq:rho}
\rho(s_i)= \begin{cases} 
				0 & \text{if    }  s_i < \beta_1\\
				v-\beta_1  &\text{if    }  \beta_1 \leq s_i \leq \beta_2 \\
				1 & \text{if    }  s_i > \beta_2,
			\end{cases}
\end{equation}
with thresholds $\beta_1$ and $\beta_2$ such that $\beta_2 - \beta_1 = 1$
and $\beta_1 < 0 < \beta_2$. 

With this energy function, the driving function $R$ would be
\begin{align}
  R(\vec s) & = \vec s - \frac{\partial E(\vec s)}{\partial \vec s}  \nonumber \\
  R_i(\vec s) & = \rho'(s_i) \left( b_i + \sum_j W_{i,j} \rho(s_j) \right).
\end{align}  
To obtain this, we have assumed that \mbox{$W_{i,j}=W_{j,i}$}. Otherwise, we would get
that \mbox{$R_i(\vec s) = \rho'(s_i) \left( b_i + \sum_j \frac{1}{2}(W_{i,j}+W_{j,i}) \rho(s_j) \right)$},
which automatically symmetrizes the weight matrix.

This formula for the driving input $R$
is similar to the usual weighted sum of firing rates, except for the novel factor $\rho'(s_i)$,
which would suggest that when a neuron is saturated (either being shut off or firing at the maximal rate),
the external inputs have no impact on its state. The only term that remains in the
neural update equation (Eq.~\ref{eq:s-update}) is the one that drives the state towards 0, i.e.,
bringing it out of the saturation region and back into a regime where the neuron is sensitive
to the outside feedback, so long as $\rho(0)$ is not a saturated value. This idea
is developed further below.

\subsubsection{Fixed Point Behavior}
\label{sec:fixed-points}

In particular, it is interesting to note what happens around a fixed point
of the state dynamics.
\begin{equation}
\label{eq:fixed-point}
 \frac{\partial E(\vec s)}{\partial \vec s}=0 \Rightarrow R(\vec s)=\vec s \Rightarrow f(\vec s,0)=\vec s
\end{equation}
which means that
\begin{equation}
  s_i = \rho'(s_i) (b_i + \sum_j W_{i,j} \rho(s_j)).
\end{equation}
Let us consider the hypothesis where the unit is saturated,
i.e., $\rho'(s_i)\approx 0$. In that case,
$R_i(\vec s)=0$ and the neural update becomes
\[
  s_{t+1,i} = (1-\epsilon) s_{t,i}
\]
which {\em converges towards} $s_i=0$.
If the origin corresponds to a region where the derivative
is significantly nonzero, $|\rho'(0)|>0$,
we get that {\bf when $\rho'(s_i)=0$, the network cannot
be at a fixed point}. Otherwise the state would move towards 0
and thus it could not have been at a fixed point.

\section{Link to Back-propagation}
\label{sec:backprop}

We are now ready to present the main result of this paper, i.e.,
a link between neural computation as inference in an energy-based
model and back-propagation of prediction error gradients.

\subsection{Propagation of Pertubations}

Consider what happens when a network such as described above sits near equilibrium,
i.e., near a fixed point as per Eq.~\ref{eq:fixed-point}. At that point, as per that
equation, the average gradient of the energy is 0 and weight updates 
are also 0 in average.

To make the link to supervised back-propagation simpler, let us consider two kinds
of visible units: input units $\vec x$ and output units $\vec y$, and $\vec s=(\vec x,\vec y,\vec h)$. Suppose that we start
by letting the network settle to a fixed point with $\vec x$ clamped to the observed
input values. Then we obtain an output $\hat{\vec y}$ at the fixed point $\hat{\vec s}$, where
$R_y(\hat{\vec s})=\hat{\vec y}$, $R_h(\hat{\vec s})=\hat{\vec h}$. Equivalently, we have
that
\begin{align}
\frac{\partial E(\hat{\vec s})}{\partial \hat{\vec y}}&=0 \nonumber \\
\frac{\partial E(\hat{\vec s})}{\partial \hat{\vec h}}&=0 
\end{align}
i.e., the ``free'' units (hidden and output) have settled to a value that
is in agreement with the clamped input units.

Now suppose that a target value $\vec y$ is observed and gradually drives the output units
from their fixed point value $\hat{\vec y}$, towards $\vec y$. This happens because the output
units are also leaky integrator neurons, meaning that their state gradually changes
based on the input they receive, in direction of the driving signal (now $\vec y$
rather than $R_y(\hat{\vec s})$.
Let us denote
\begin{equation}
  \Delta \vec y=\epsilon(\vec y-\hat{\vec y})
  \end{equation}
that 
initial change of $\hat{\vec y}$ when going from time step 0 to time step 1
(following the neural update equation~\ref{eq:s-update}, 
Let us consider as training objective the squared prediction error
but with  $R_y(\hat{\vec s})$ replaced by $\vec y$).
That would push the global state $\hat{\vec s}$ away from the equilibrium where it was sitting,
and into a region where $\frac{\partial E(\vec s)}{\partial \vec s}$ is non-zero. 
\begin{equation}
  C = \frac{1}{2}||\hat{\vec y}-\vec y||^2,
\end{equation}
which corresponding to the mismatch between the prediction $R_y(\hat{\vec s})$ and the target
value $\vec y$ driving the output units, or equivalently
\begin{equation}
\label{eq:C-bp}
  C = \frac{1}{2}||R_y(\hat{\vec s}) - \vec y||^2 = \frac{1}{2\epsilon^2}||\Delta \vec y||^2,
\end{equation}
because at equilibrium $R_y(\hat{\vec s})=\hat{\vec y}$.
Note that 
\begin{equation}
  \Delta \vec y = - \epsilon \frac{\partial C}{\partial \hat{\vec y}},
\end{equation}
where $\epsilon$ can be seen as a learning rate if we were trying to do SGD on $\hat{\vec y}$ directly.

Now, how would the rest of the network react to this external perturbation?
Each hidden neuron would approximately move in the direction of the gradient
of $C$, but only those (call them $\vec h_1$) that are directly connected to the output would initially
feel the pressure to minimize $C$. That perturbation (in the form of a volley of additional spikes,
for real neurons) would then travel to the next circle of neurons, those directly connected to $\vec h_1$
but not to the output, etc.

Let us look at this in more detail. Consider a typical multi-layer architecture with connections
between the output layer (yielding value $\hat{\vec y}$ at equilibrium)
and the top hidden layer (yielding value $\hat{\vec h}_1$), between
the top  and the next hidden layer (yielding $\hat{\vec h}_1$ and $\hat{\vec h}_2$), etc.
The change $\Delta \vec y$ would propagate to $\hat{\vec h}_1$ via the neural update, which, when
we ignore the effect of the injected noise, would yield a change in $\hat{\vec h}_1$
\begin{equation}
  \Delta \vec h_1 = \epsilon (R_{h_1}((\vec x,\hat{\vec y}+\Delta \vec y, \hat{\vec h})) - \hat{\vec h}_1)\enspace.
\end{equation}
With $\Delta \vec y$ small (arising out of our assumption that the visible units only gradually
move towards their target), we can approximate the above by taking the Taylor expansion of $R$ around $\hat{\vec y}$, 
\begin{equation}
  R_{h_1}(( \vec x,\hat{\vec y}+\Delta \vec y, \hat{\vec h})) = \hat{\vec h}_1 + \frac{\partial R_{ h_1}(\hat{\vec s})}{\partial \hat{\vec y}} \Delta \vec y + o(\epsilon) \enspace.
\end{equation}
exploiting $R_{h_1}(\hat{\vec s})=\hat{\vec h}_1$ at the fixed point, yielding
\begin{equation}
  \Delta \vec h_1 = \epsilon \frac{\partial R_{h_1}(\hat{\vec s})}{\partial \hat{\vec y}} \Delta \vec y + o(\epsilon^2)
\end{equation}
Hence we have that
\begin{equation}
\label{eq:delta-h1}
  \Delta \vec h_1 = - \epsilon^2 \frac{\partial R_{ h_1}(\hat{\vec s})}{\partial \hat{\vec y}} \frac{\partial C}{\partial \hat{\vec y}} + o(\epsilon^2)\enspace.
\end{equation}
  Note, that with the assumed layer-wise neural network structure
  (i.e. having no connections between neurons insight one layer)
  $o(\epsilon)$ (and $o(\epsilon^2)$ respectively) is zero, if the neural
  non-linearity $\rho$ fulfills $\rho''(s_i)=0$, which is the case for the
  non-linearity  given in Eqs. \ref{eq:rho}. This is a consequence of having all the higher derivatives
  of $\rho$ being zero, since $\rho$ is piecewise linear.

In order to obtain backprop, what we would like to get, though is not $\frac{\partial R_{h_1}(\hat{\vec s})}{\partial \hat{\vec y}} \Delta \vec y$
but $\frac{\partial R_y(\hat{\vec s})}{\partial \hat{\vec h}_1} \Delta \vec y$ since that would correspond to an application of the
chain rule and we would have $\Delta \vec h_1 \propto \frac{\partial C}{\partial \hat{\vec h}_1}$. The good news
is that this equality is true because $R$ is a first derivative of a function related to the energy function:
\begin{equation}
  R(\vec s) = \vec s -\frac{\partial E(\vec s)}{\partial \vec s} = \frac{\partial L(\vec s)}{\partial \vec s}
\end{equation}
where 
\begin{equation}
  L(\vec s) = \frac{1}{2} ||\vec s||^2 - E(\vec s) 
\end{equation}
and $R_y(\hat{\vec s})=\frac{\partial L(\hat{\vec s})}{\partial \hat{\vec y}}$, $R_{h_1}(\hat{\vec s})=\frac{\partial L(\hat{\vec s})}{\partial \hat{\vec h}_1}$,
so that $\frac{\partial R_{h_1}(\hat{\vec s})}{\partial \hat{\vec y}}$
and $\frac{\partial R_y(\hat{\vec s})}{\partial \hat{\vec h}_1}$ are cross-derivatives of $L$.
As we know that cross derivatives are symmetric, 
\begin{equation}
  \frac{\partial R_{h_1}(\hat{\vec s})}{\partial \hat{\vec y}} = \frac{\partial^2 L}{\partial \hat{\vec y} \partial \hat{\vec h}_1} = 
  \left(\frac{\partial^2 L}{\partial \hat{\vec h}_1 \partial \hat{\vec y}}\right)^T = \frac{\partial R_y(\hat{\vec s})}{\partial \hat{\vec h}_1}^T.
\end{equation}
Now note that since at the fixed point $\hat{\vec y}=R_y(\hat{\vec s})$
\begin{equation}
  \frac{\partial R_y(\hat{\vec s})}{\partial \hat{\vec h}_1} = \frac{\partial \hat{\vec y}}{\partial \hat{\vec h}_1}
\end{equation}
and we are ready to exploit that to rewrite $\Delta \vec h_1$ (Eq.~\ref{eq:delta-h1}) in a form that
equates it with a backpropagated gradient:
\begin{align}
\label{eq:delta-h1-delta-x}
  \Delta \vec h_1 & = - \epsilon^2 \frac{\partial R_y(\hat{\vec s})}{\partial \hat{\vec h}_1}^T \frac{\partial C}{\partial \hat{\vec y}} + o(\epsilon^2) \nonumber \\
             & = - \epsilon^2 \frac{\partial \hat{\vec y}}{\partial \hat{\vec h}_1}^T \frac{\partial C}{\partial \hat{\vec y}} + o(\epsilon^2) \nonumber \\
             & = -  \epsilon^2\frac{\partial C}{\partial \hat{\vec h}_1} + o(\epsilon^2) \nonumber \\
\end{align}
Similarly, the perturbation $\Delta \vec h_1$ will be transmitted at the next time step
to the units $\hat{\vec h}_2$ that are directly connected to $\hat{\vec h}_1$ (but not to $\hat{\vec y}$), and yield
\begin{align}
  \Delta \vec h_2 & = \epsilon \frac{\partial \hat{\vec h}_1}{\partial \hat{\vec h}_2}^T \Delta \vec h_1 + o(\epsilon^3) \nonumber \\
  \Delta \vec h_2 & = - \epsilon^3 \frac{\partial \hat{\vec h}_1}{\partial \hat{\vec h}_2}^T \frac{\partial C}{\partial \hat{\vec h}_1} + o(\epsilon^3) \nonumber \\
  \Delta \vec h_2 & = - \epsilon^3 \frac{\partial C}{\partial \hat{\vec h}_2} + o(\epsilon^3).
\end{align}

\subsection{Stochastic Gradient Descent Weight Update}

The above result is consistent with and inspired by
the idea previously proposed by~\citet{Hinton-DL2007}
that temporal change can encode back-propagated gradients.
What would it take for the $\Delta \hat{\vec h}_k$ at layer $k$ to turn into a stochastic
gradient descent (SGD) weight update with respect to the prediction error $||\vec y-\hat{\vec y}||^2$?
Since the
state change $\dot{\vec s}$ represents the gradient of the prediction error with respect to $\vec s$,
SGD on $W_{i,j}$ would require
the weight change $\Delta W_{i,j}$ being proportional to the rate of change of the state of the
post-synaptic neuron, $\dot{s}_i$ and proportional to the gradient of the fixed point state $\vec s$ 
with respect to $W_{i,j}$, i.e., to $\frac{\partial s_i}{\partial W_{i,j}}$. To first approximation, this is
the firing rate of the pre-synaptic neuron, $\rho(s_j)$. This would yield
\begin{equation}
  \label{eq:delta-w-stdp}
  \Delta W_{i,j} \propto \dot{s}_i \rho(s_j) \enspace.
\end{equation}
It turns out that such a learning rule allows to simulate the relationship between spike-timing
and synaptic change according to the STDP (spike-timing dependent plasticity), as
shown via simulations by~\citet{Bengio-et-al-arxiv2015b}. See~\citet{Xie+Seung-NIPS1999}
for a similar learning rule, also related to STDP via a different analysis.

Thus, with this STDP-compatible learning rule,
the change in weights due to the initial perturbation would be approximately
proportional to the back-propagation update, since
it corresponds to $\Delta \vec h \frac{\partial\vec  h}{\partial W}$. However, note the multiplicative factors $\epsilon^{k+2}$
for units $\hat{\vec h}_k$ at layer $k$, that make the initial changes much slower for the more remote layers.
This is because the leaky integration neurons have not had time to integrate the information yet,
so practically it will take on the order of the time constant times $k$ for the change in $\hat{\vec h}_k$
to become significant, unless we adjust the per-layer learning rates accordingly.

Although we see that the proposed neural dynamics and weight updates will behave approximately
like back-propagation, there are differences, especially when we consider what happens after more
time steps. But maybe the most important take-home message from this link with back-propagation
is the following. We know that back-propagation works extremely well to train both supervised
and unsupervised networks. We see here that back-propagation essentially corresponds to a variational
update when the inference is infinitesimal, i.e., we only allow a single step of inference
corresponding to small moves in the direction of reducing the energy function.

\sb
\section{Related work, contributions and future work}
\sb

An important inspiration for this work is the idea proposed by~\citet{Hinton-DL2007}
that brains could implement back-propagation by using temporal derivatives
to represent activation gradients, and the suggestion that combining this
assumption with STDP would approximately yield SGD on the synaptic weights.

The idea of neural computation corresponding to a form of stochastic
relaxation towards lower energy configurations is of course very old, for example
with the Boltzmann machine~\citep{Hinton-bo86} and its Gibbs sampling
procedure.  For more recent work in this direction, see also~\citep{Berkes-et-al-2011}.
What differs here from the 
Boltzmann machine is that we consider the state space to be
continuous (associated with the expected voltage potential, integrating out
the random effects due to spikes), rather than discrete, and that we
consider very small steps (going down the gradient of the energy), which is
more like a Langevin MCMC, rather than allowing each neuron to stochastically jump
with higher probabibility to its
optimal state, given its neighbors configuration, which is
what Gibbs sampling does.

There are of course many other papers on theoretical interpretations of STDP,
and the reader can find many references in~\citet{Markram-et-al-2012}, but more 
work is needed to explore the connection of STDP to an unsupervised learning objective that
could be used to train not just a single layer network (like PCA and
traditional Hebbian updates) but also a deep unsupervised model. Many approaches
~\citep{Fiete+Seung-2006,Rezende+Gerstner-2014} rely on variants of the REINFORCE 
algorithm~\citep{Williams-1992}
to estimate the gradient of a global objective function (basically by correlating
stochastic variations at each neuron with the changes in the global objective).
Although this principle is simple, it is not clear that it will scale to very
large networks due to the linear growth of the variance of the estimator with the number
of neurons. It is therefore tempting to explore other avenues, and we hope
that the building blocks introduced here and the links made with energy-based approaches
with variational inference for 
unsupervised learning can form useful material for a more efficient unsupervised learning
principle for deep networks that is also consistent with STDP.

If the energy function proposed here is closer to a biological truth than the energy defined
for continuous Hopfield networks~\citep{Hopfield84}, we should see that (a) firing rate returns to its
baseline when the neuron is saturated (completely turned off or maximally turned on) and (b)
synaptic weight changes should also vanish under this condition, as seen by inspection of Eq.~\ref{eq:delta-w-stdp}.
It would clearly be interesting to test these predictions in actual biological experiments.

Much remains to be done to obtain a complete probabilistic theory of unsupervised learning
that is consistent with STDP, but we believe that we have put interesting ingredients in place.
One aspect that requires a lot more development is how the proposed STDP update
helps to fit the sensory observations $\vec x$. 
If, as hypothesized above, neural computation is approximately doing inference (e.g. Langevin MCMC),
then each step of inference, in average, brings us towards an equally likely or
even more likely configuration of $\vec h$, given $\vec x$, according to the model. Hence each step is
approximately pointing down the energy of $P(\vec h|\vec x)$. Now, in an EM or variational EM 
context such as discussed in~\citet{emview,Kingma+Welling-ICLR2014,Bengio-et-al-arxiv2015}, with $x$ fixed,
the distribution we want to model and consider as a target towards which parameters
should be updated is precisely the joint of $h \sim P(\vec h|\vec x)$ and $\vec x \sim$ the training data,
which we now call $Q(\vec h,\vec x)$ (the inference distribution), following the above papers.
We would like the parameters to move in the direction that makes the model more consistent with $Q(\vec h,\vec x)$,
which is what is required to maximize the variational EM bound on the data likelihood $P(\vec x)$.
The idea is that we change the inference process so that it would reach its final state
faster, which corresponds to a configuration of $\vec h$ that fits well the observed $\vec x$.

Another open question is how to reconcile the need for symmetric weights when
we introduce an energy function and the fact that $W_{i,j}$ and $W_{j,i}$ are stored
at two physically different places in biological neurons. An encouraging observation
is that earlier work on
auto-encoders empirically showed that even when the forward and backward weights
are not tied, they tend to converge to symmetric values, and in the linear case
the minimization of reconstruction error automatically yields symmetric weights~\citep{Vincent-JMLR-2010-small}.
Another encouraging piece of evidence, also linked to autoencoders, is the
theoretical result from~\citet{Arora-et-al-2015}, showing that the symmetric solution
minimizes the autoencoder reconstruction error between two successive layers of
rectifying (ReLU) units.

\ifarxiv
\section*{Acknowledgments} 
 
The authors would like to thank Benjamin Scellier, Thomas Mesnard, Saizheng Zhang, Yuhuai Wu,
Dong-Hyun Lee, Jyri Kivinen, Jorg Bornschein, Roland Memisevic and Tim Lillicrap
 for feedback and discussions, as well as NSERC, CIFAR, Samsung
and Canada Research Chairs for funding.
\else
\fi

\bibliography{strings,ml}

\begin{thebibliography}{}

\bibitem[Alain and Bengio(2013)Alain and Bengio]{Alain+Bengio-ICLR2013-small}
Alain, G. and Bengio, Y. (2013).
\newblock What regularized auto-encoders learn from the data generating
  distribution.
\newblock In {\em ICLR'2013\/}. also arXiv report 1211.4246.

\bibitem[Andrieu {\em et~al.}(2003)Andrieu, de~Freitas, Doucet, and
  Jordan]{Andrieu2003}
Andrieu, C., de~Freitas, N., Doucet, A., and Jordan, M. (2003).
\newblock An introduction to {MCMC} for machine learning.
\newblock {\em Machine Learning\/}, {\bf 50}, 5--43.

\bibitem[Arora {\em et~al.}(2015)Arora, Liang, and Ma]{Arora-et-al-2015}
Arora, S., Liang, Y., and Ma, T. (2015).
\newblock Why are deep nets reversible: a simple theory, with implications for
  training.
\newblock Technical report, arXiv:1511.05653.

\bibitem[Bengio {\em et~al.}(2015a)Bengio, Mesnard, Fischer, Zhang, and
  Wu]{Bengio-et-al-arxiv2015b}
Bengio, Y., Mesnard, T., Fischer, A., Zhang, S., and Wu, Y. (2015a).
\newblock {STDP} as presynaptic times rate of change of postsynaptic activity.
\newblock arXiv:1509.05936.

\bibitem[Bengio {\em et~al.}(2015b)Bengio, Lee, Bornschein, and
  Lin]{Bengio-et-al-arxiv2015}
Bengio, Y., Lee, D.-H., Bornschein, J., and Lin, Z. (2015b).
\newblock Towards biologically plausible deep learning.
\newblock arXiv:1502.04156.

\bibitem[Berkes {\em et~al.}(2011)Berkes, Orban, Lengyel, and
  Fiser]{Berkes-et-al-2011}
Berkes, P., Orban, G., Lengyel, M., and Fiser, J. (2011).
\newblock Spontaneous cortical activity reveals hallmarks of an optimal
  internal model of the environment.
\newblock {\em Science\/}, {\bf 331}, 83--–87.

\bibitem[Fiete and Seung(2006)Fiete and Seung]{Fiete+Seung-2006}
Fiete, I.~R. and Seung, H.~S. (2006).
\newblock Gradient learning in spiking neural networks by dynamic perturbations
  of conductances.
\newblock {\em Physical Review Letters\/}, {\bf 97}(4).

\bibitem[Friston and Stephan(2007)Friston and Stephan]{Friston+Stephan-2007}
Friston, K.~J. and Stephan, K.~E. (2007).
\newblock Free-energy and the brain.
\newblock {\em Synthese\/}, {\bf 159}, 417--–458.

\bibitem[Hinton(2007)Hinton]{Hinton-DL2007}
Hinton, G.~E. (2007).
\newblock How to do backpropagation in a brain.
\newblock Invited talk at the NIPS'2007 Deep Learning Workshop.

\bibitem[Hinton and Sejnowski(1986)Hinton and Sejnowski]{Hinton-bo86}
Hinton, G.~E. and Sejnowski, T.~J. (1986).
\newblock Learning and relearning in {Boltzmann} machines.
\newblock In D.~E. Rumelhart and J.~L. McClelland, editors, {\em Parallel
  Distributed Processing: Explorations in the Microstructure of Cognition.
  Volume 1: Foundations\/}, pages 282--317. MIT Press, Cambridge, MA.

\bibitem[Hopfield(1984)Hopfield]{Hopfield84}
Hopfield, J.~J. (1984).
\newblock Neurons with graded responses have collective computational
  properties like those of two-state neurons.
\newblock {\em Proceedings of the National Academy of Sciences, USA\/}, {\bf
  81}.

\bibitem[Kingma and Welling(2014)Kingma and Welling]{Kingma+Welling-ICLR2014}
Kingma, D.~P. and Welling, M. (2014).
\newblock Auto-encoding variational bayes.
\newblock In {\em Proceedings of the International Conference on Learning
  Representations (ICLR)\/}.

\bibitem[Markram {\em et~al.}(2012)Markram, Gerstner, and
  Sjöström]{Markram-et-al-2012}
Markram, H., Gerstner, W., and Sjöström, P. (2012).
\newblock Spike-timing-dependent plasticity: A comprehensive overview.
\newblock {\em Frontiers in synaptic plasticity\/}, {\bf 4}(2).

\bibitem[Neal and Hinton(1999)Neal and Hinton]{emview}
Neal, R. and Hinton, G. (1999).
\newblock A view of the {EM} algorithm that justifies incremental, sparse, and
  other variants.
\newblock In M.~I. Jordan, editor, {\em Learning in Graphical Models\/}. MIT
  Press, Cambridge, MA.

\bibitem[Rezende and Gerstner(2014)Rezende and Gerstner]{Rezende+Gerstner-2014}
Rezende, D.~J. and Gerstner, W. (2014).
\newblock Stochastic variational learning in recurrent spiking networks.
\newblock {\em Frontiers in Computational Neuroscience\/}, {\bf 8}(38).

\bibitem[Vincent(2011)Vincent]{Vincent-2011}
Vincent, P. (2011).
\newblock A connection between score matching and denoising autoencoders.
\newblock {\em Neural Computation\/}, {\bf 23}(7).

\bibitem[Vincent {\em et~al.}(2010)Vincent, Larochelle, Lajoie, Bengio, and
  Manzagol]{Vincent-JMLR-2010-small}
Vincent, P., Larochelle, H., Lajoie, I., Bengio, Y., and Manzagol, P.-A.
  (2010).
\newblock Stacked denoising autoencoders: Learning useful representations in a
  deep network with a local denoising criterion.
\newblock {\em J. Machine Learning Res.}, {\bf 11}.

\bibitem[Williams(1992)Williams]{Williams-1992}
Williams, R.~J. (1992).
\newblock Simple statistical gradient-following algorithms connectionist
  reinforcement learning.
\newblock {\em Machine Learning\/}, {\bf 8}, 229--256.

\bibitem[Xie and Seung(2000)Xie and Seung]{Xie+Seung-NIPS1999}
Xie, X. and Seung, H.~S. (2000).
\newblock Spike-based learning rules and stabilization of persistent neural
  activity.
\newblock In S.~Solla, T.~Leen, and K.~M\"{u}ller, editors, {\em Advances in
  Neural Information Processing Systems 12\/}, pages 199--208. MIT Press.

\end{thebibliography}

\bibliographystyle{natbib}

\end{document}